\def\year{2022}\relax
\documentclass[letterpaper]{article} % DO NOT CHANGE THIS
\usepackage{aaai22}  % DO NOT CHANGE THIS
\usepackage{times}  % DO NOT CHANGE THIS
\usepackage{helvet}  % DO NOT CHANGE THIS
\usepackage{float}
\usepackage{courier}  % DO NOT CHANGE THIS
\usepackage[hyphens]{url}  % DO NOT CHANGE THIS
\usepackage{graphicx} % DO NOT CHANGE THIS
\usepackage[numbers]{natbib}  % DO NOT CHANGE THIS AND DO NOT ADD ANY OPTIONS TO IT
\usepackage{caption} % DO NOT CHANGE THIS AND DO NOT ADD ANY OPTIONS TO IT
\DeclareCaptionStyle{ruled}{labelfont=normalfont,labelsep=colon,strut=off} % DO NOT CHANGE THIS
\usepackage{algorithm}
\usepackage{algorithmic}
\usepackage{xcolor}

\usepackage{newfloat}
\usepackage{listings}
\floatstyle{ruled}
\newfloat{listing}{tb}{lst}{}
\floatname{listing}{Listing}
\usepackage{ amssymb }
\usepackage{fancyhdr}
\title{FREDSR: Fourier Residual Efficient Diffusive GAN for Fast Single Image Super Resolution

    }
\author {
    Kyoungwan Woo\textsuperscript{\rm 1 *}
    Achyuta Rajaram \textsuperscript{\rm 2 *}
}
\affiliations {
    \textsuperscript{\rm 1} Massachusetts Institute of Technology\\
    \textsuperscript{\rm 2} Phillips Exeter Academy\\
    kwoo1@mit.edu, achyuta.rajaram@gmail.com\\
    \centering
    \includegraphics[width=0.95\textwidth]{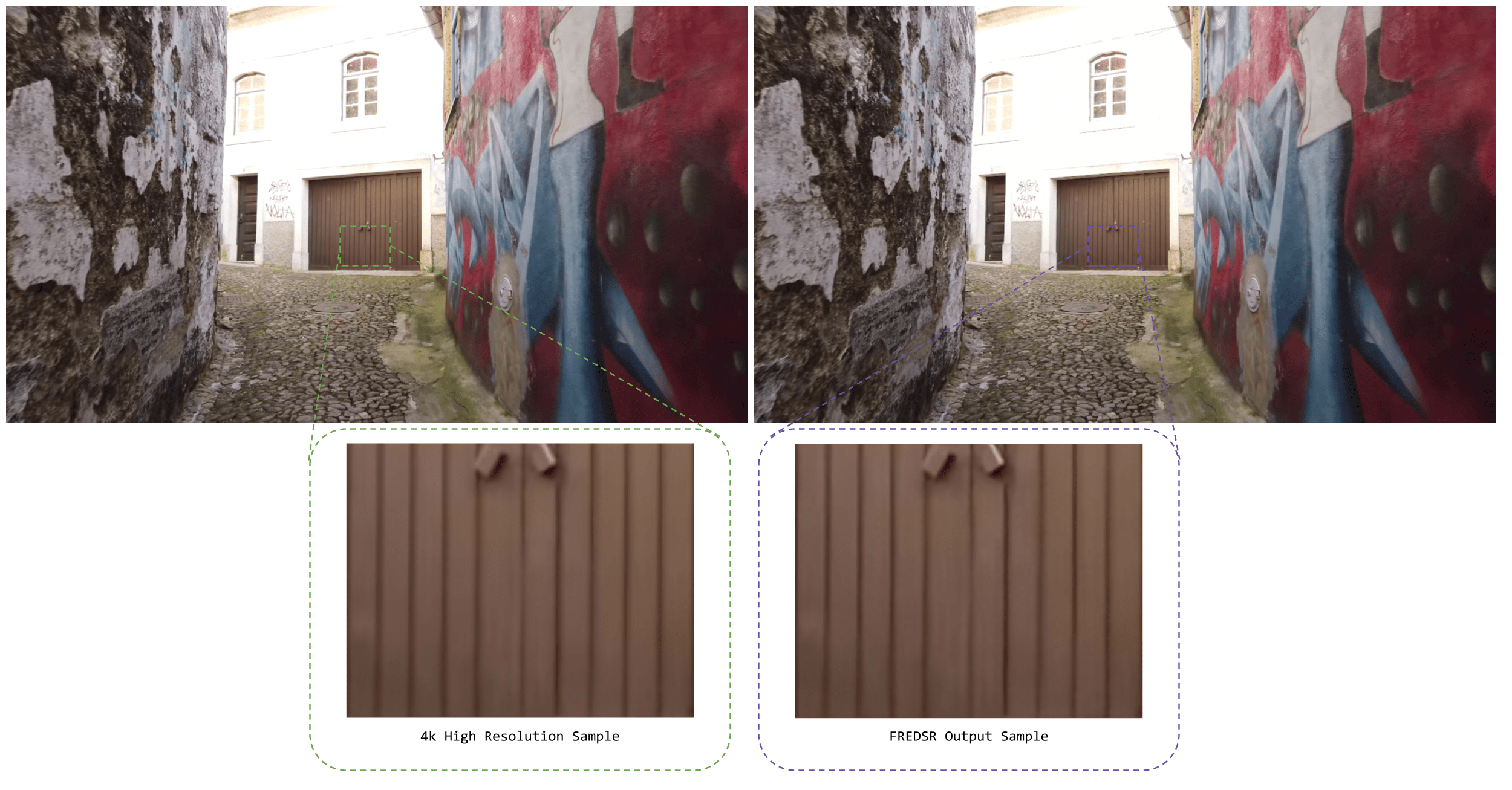}
    \onecolumn
    \raggedright
    Figure 1: Shown above are the original UHDSR4K images compared to the up-scaled output images of FREDSR, left to right. Note that on repeated patterns, such as the door given, we reach near identical reproduction, even on extremely high resolution. We can  attribute this to the high perceptive field of our model, .
    \twocolumn
    
}

\begin{document}

\maketitle

\begin{abstract}
\emph{
FREDSR is a GAN variant that aims to outperform traditional GAN models in specific tasks such as Single Image Super Resolution with extreme parameter efficiency at the cost of per-dataset generalizeability. FREDSR integrates fast Fourier transformation, residual prediction, diffusive discriminators, etc to achieve strong performance in comparisons to other models on the UHDSR4K dataset for Single Image 3x Super Resolution from 360p and 720p with only 37000 parameters. The model follows the characteristics of the given dataset, resulting in lower generalizeability but higher performance on tasks such as real time up-scaling.}
\end{abstract}
\section{Introduction}
\hspace{\parindent} Image to Image Translation(I2I) is a highly general meta-task that can summarize most application-specific subtasks in image generation or processing. Broadly, we use the definition from \cite{2101.08629}, where we seek to convert an an input image from domain A to domain B, while preserving "intrinsic source content", i.e. the meaningful semantic features of the image, while converting only the "extrinsic style", pertaining to the specific domain. This process can be defined as a mapping that synthesizes images that are indistinguishable from a sample from the target distribution B, using the semantic information present from a given sample of distribution A. This definition can generalize to many problems in image processing, with past work being used for image synthesis \cite{Zhu_2020_CVPR}, image segmentation \cite{GUO2020127}, style transfer \cite{8237506}, image inpainting \cite{DBLP:journals/corr/abs-2109-07161}, image super-resolution \cite{8575264}, among many other applications. \par Given the formulation of the I2I problem stated above, it seems natural that Generative Adversarial Networks(GAN) would be applicable for these tasks. GANs seek to generate images indistinguishable from those in a given distribution by playing a mini-max game between a discriminator and generator, where the generator seeks to decrease the discriminator's accuracy in determining if the outputs of the generator belong to a specified distribution of images, while the discriminator seeks to maximise said accuracy \cite{goodfellow}. This methodology circumvents the problems of using a pixel-level loss, or other naive supervised losses, as when they are used, blurry images are creates due to averaging of the multiple plausible solutions for a given translation. \par As a result of this natural applicability to GANs, past research has sought to apply GAN architectures to I2I translation problems, such as the pix2pix network, which pioneered the development of GANs for I2I translation \cite{isola}. Pix2pix allowed for high-quality low-resolution I2I translation to be computed without the need for problem-specific hand crafted loss functions. However, the initial formulation has not been crafted for high-resolutions, with \cite{chen} stating that conditional GANs such as pix2pix can struggle to generate high definition images, due to training instability. Furthermore, they indicate that including Perceptual Loss, an error function based on the outputs of classification networks, that seek to perform large-scale feature extraction, has significant benefits. Using this as a supervised objective allowed for stable images and was an improvement over raw pixel-level objectives.  \par More recently, an improvement to pix2pix has been suggested, referred to as pix2pixHD, combining classifier-based perceptual loss with discriminator based loss in order to allow for high resolution image generation \cite{wang2018}. However, this model requires 3 discriminators to allow for the processing of information at every scale, which creates large increases to the computational resources required during training. Furthermore, using several discriminators increases learning instability, as matching gradient updates between the Generator and Discriminator is one of the largest challenges in GAN training, as not doing so properly leads to common GAN failure modes such as Catastrophic Forgetting and Mode Collapse \cite{thantung}.\par An alternative to this architecture, optimized for large mask high resolution image inpainting, uses Fast Fourier Convolutions (FFCs) to allow for efficient processing of global-scale information along with the local-scale details \cite{DBLP:journals/corr/abs-2109-07161},\cite{fastfourier}. With this prior, we contribute:
\par\textbf{\textit{i.}} Using the previously introduced FFC block, we create a novel model architecture for I2I translation, adapting the original encoder-decoder stack from pix2pix. The use of the FFC allows the model to quickly gain access to global information. This allows for the model to operate on extrinsic global information throughout its entirety, improving its parameter efficiency for translation tasks. Additionally, this architecture is easily adaptable for Single Image Super Resolution(SISR) problems, and the maintaining of global information proves crucial to the performance of this architecture.
\par\textbf{\textit{ii.}}
We extend the diffusive discriminator from \cite{wang2022}, which seeks to combat discriminator overfitting by using a tool from the highly successful de-noising Diffusion Probabilistic Model architecture \cite{ho2020}, adding diffusion from a Gaussian mixture distribution to the discriminator inputs, by using a residual discriminator design.
\par Overall, we use all of these techniques to create a novel GAN variant, which experimentation has shown to be performant for specialized SISR tasks, while having much higher parameter efficiency in orders of magnitude. We motivate the existence of such a technique in the application of real-time 3D rendering, such as those found in modern video games. Modern gaming requires high resolution rendering of complicated 3D meshes with potentially billions of polygons, as well as expensive processes such as ray-traced per pixel lighting, ambient occlusion, and motion blur. Performance in this application has traditionally scaled harshly with output resolution, due to a quadratic increase in total calculations. Thus, a real-time application of SISR can provide efficiency benefits. One of the examples of utilizing unsupervised learning techniques for this task was introduced by NVIDIA with their Deep Learning Super Sampling application. FREDSR follows a similar notion of extremely performant, yet specialized SISR problem task, and was optimized for high performance on a specific dataset with extreme parameter efficiency.

\section{Method}
\hspace{\parindent} We seek to develop a novel Single Image Super  Resolution(SISR) architecture where we take a three channel input image high resolution and low resolution pair $q_{hr}$ and $q_{lr}$. We propose a model architecture that takes in $q_{lr}$ and outputs a  three channel image $q_h$, which has the same resolution as $q_{hr}$.
\begin{figure*}[!ht]
    \centering
    \includegraphics[width=\textwidth]{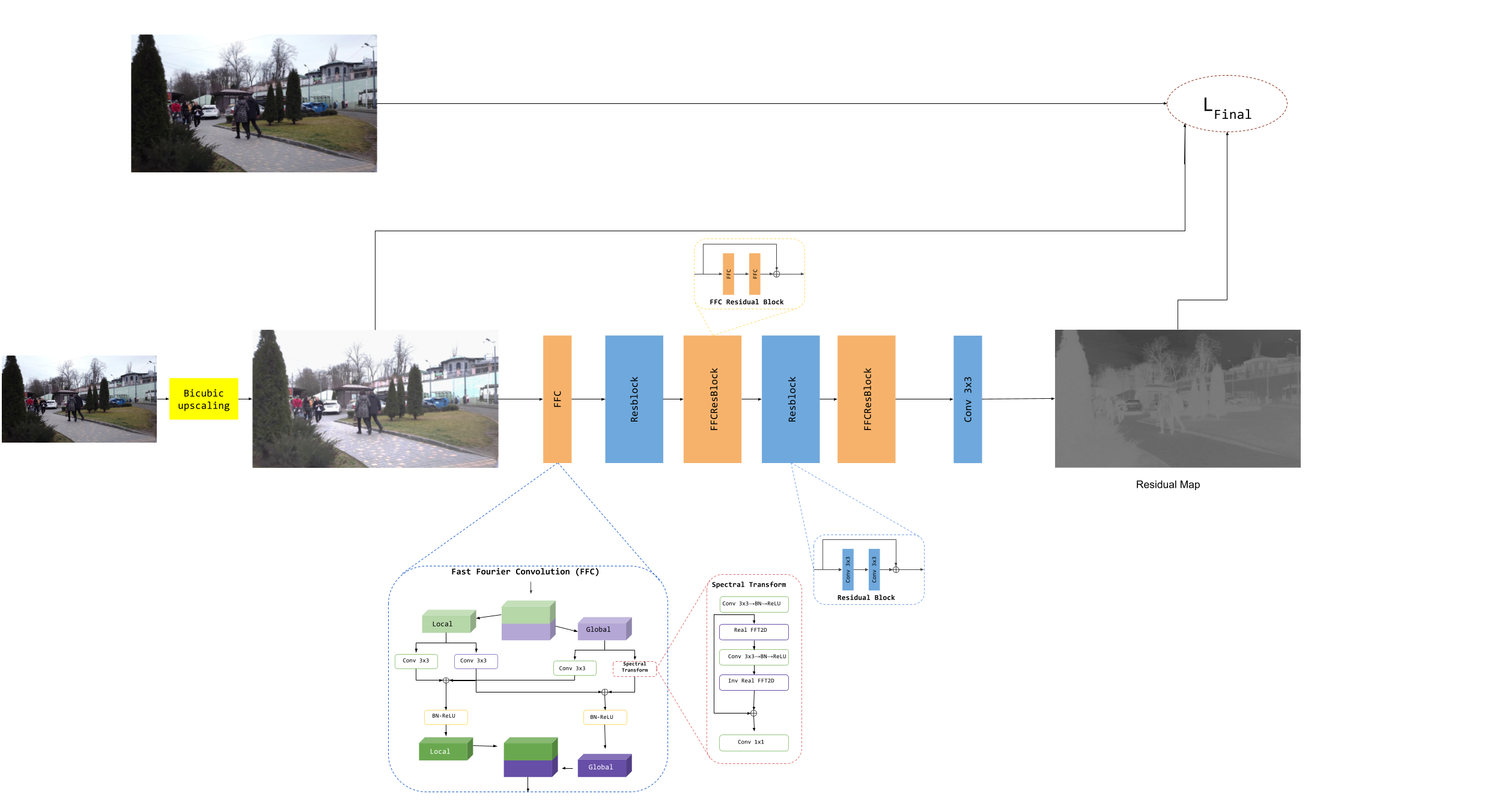}
    \caption*{Figure 2: The model architecture of the FREDSR generator. This network uses the recently proposed FFCs \cite{fastfourier}, along with a complex loss function that combines adversarial, perceptual, and pixel-level losses. Graphic style was inspired by \cite{DBLP:journals/corr/abs-2109-07161}.}
\end{figure*}
\subsection{Maintaining global information through all layers}
\hspace{\parindent} Challenging problems in I2I translation, including that of large-scale modification to high-resolution data, requires processing of global information. We argue that for successful high-resolution I2I translation, a high receptive field that encapsulates global information, and maintains it throughout the network is necessary in order to efficiently process global extrinsic parameters on an image, such as the time of day, lighting conditions, the artistic style associated with a painter, etc. Traditional fully convolutional models suffer from slow growth of effective receptive field, due to the small  kernels used \cite{NIPS2016_c8067ad1}. To improve on this, \cite{yu2016} proposed the dilated convolution, which introduces "gaps" between the values of a convolution, "spreading out" a 3x3 convolution, for example over an 8x8 area, while only using 9 parameters. However, this loses some local information present, meaning that a complex architecture that combines these with traditional convolutions is necessary in order to achieve usable results, as was seen in the I2I density map generation network proposed in \cite{li2018}. Additionally, even the introduction of such dilated convolutions only slightly increases the receptive field. Thus, as a result, improvement is possible, which comes by way of the \textbf{Fast Fourier Convolution (FFC)}\cite{fastfourier}, based on the Fast Fourier Transform \cite{brigham}. \newline\newline FFC splits information into two main channels, global and local, where the local information is processed using traditional convolutions, and the global information is processed using a real FFT2d. During each FCC block, the information of both branches are combined to allow for more efficient use of parameters. An illustration of the exact construction of the FFC block can be seen in Figure 2. This allows for the efficient use of parameters. This unit are fully differentiable, and we show them in used as a drop-in replacement for traditional convolutions. Additionally, FFCs have been showed to be well suited to capture periodic structures, which are extremely common both in natural and artificial environments, such as brick buildings or waves on the surface of water \cite{DBLP:journals/corr/abs-2109-07161}.

\subsection{Model Architecture}
Here, we introduce our novel I2I architecture, and the modifications necessary for high resolution SISR tasks with an extremely efficient network. We have already described the foundational units of this architecture, the FFC block. This architecture is illustrated in Figure 2. 
\subsubsection{SISR Generator} 
A traditional I2I network cannot perform SISR, due to the differing resolutions of the input and output. However, our architecture is easily adaptable to perform this task. We perform this adaption based off of the foundation of the VDSR model proposed in \cite{Kim2015}. to begin, we upscale the low-resolution input by using a bicubic upscaler. Then we use our model to predict the difference in value between the input bicubic upscaled image, and the true image. This process is visualized in figure 2. This model has several advantages over other super-resolution networks, as the FFC module allow for the more efficient use of parameters, preservation of low-frequency features maintained through the upscaling. Additionally the residual learning allows for much more efficient use of parameters and faster training, as no training time or parameters are "wasted" learning to copy the original image. This allows our model to achieve competitive performance to SOTA methods with with only 37115 total parameters on the specialized tasks.
\begin{figure}[!ht]
    \centering
    \includegraphics[width=0.47\textwidth]{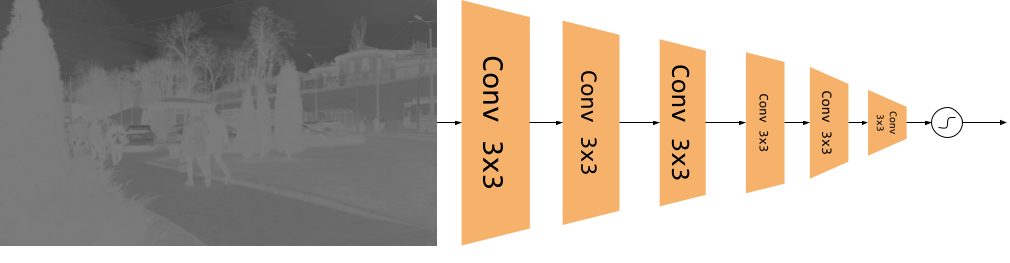}
    \caption*{Figure 3: The model architecture of the FREDSR discriminator. This was based on a traditional discriminator, with special modifications for the super-resolution task. This discriminator uses diffusion applied to its inputs, among other tactics to improve discriminator performance and prevent overfitting. Graphic style was inspired by \cite{DBLP:journals/corr/abs-2109-07161}.}
\end{figure}
\subsubsection{SISR Discriminator}
We allow the discriminator to train to distinguish between the residuals, also known as the differences between the upscaled images and the true images, and our model's predicted residuals. This architecture is visualized in full in figure 3.
 
\subsection{Generative Loss Functions}
In FREDSR, we combine various L1 and L2 losses. The benefits of combining various level losses have been previously discussed in papers such as \cite{isola}.
\subsubsection{SSIM loss}
In the case of L2 loss, we utilize the Structural Similarity Index Metric for the SISR model. In conjunction with the residual discriminator, SSIM works to deblurr the image. Compared to common L1/L2 losses such as MAE and MSE, SSIM has been proved to outperform in super resolution tasks, such as in \cite{zhao2018}. The loss function is as follows:
\begin{equation}
  \mathcal{L}_{SSIM}= -SSIM(x,y) = -\frac{(2\mu_x\mu_y + C_1) + (2 \sigma _{xy} + C_2)} 
    {(\mu_x^2 + \mu_y^2+C_1) (\sigma_x^2 + \sigma_y^2+C_2)}
  \label{eq:SSIM}
\end{equation}
With x,y defined as the generated and real images respectively.
\subsubsection{Charbonnier Loss}
As a method of enforcing color similarity, as well as mixing L1 and L2 losses, we use the Charbonnier Loss function. This is a variation of MAE which is differentiable and has shown to mix beneficial properties of L1 and L2 losses, and has shown significant success.
\begin{equation}
  \mathcal{L}_{CHARB} = E(\sqrt{(x-y)^2 + \epsilon})
  \label{eq:MGE}
\end{equation}
With x,y defined as the generated and real images respectively, and $\epsilon$ a constant chosen.
\subsubsection{MGE loss}
To improve the performance of the model for SISR tasks, we add an additional loss on top of the global feature reconstruction loss. This loss is focused on the maintaining of edges, and was introduced for SISR in \cite{lu2019}. The maintaining of sharp edges allows for a more accurate image to that captured at high resolution. MGE uses a classical sobel operator, first introduced in \cite{kanopoulos}. After this, we can  compute the gradients for each pixel, then define MGE as follows.
\begin{equation}
  \mathcal{L}_{MGE} = E((G(x)-G(y))^2)
  \label{eq:MGE}
\end{equation}
With x,y defined as the generated and real images respectively.

\subsubsection{Adversarial Loss}
In the case of adversarial losses, as the model implementation follows a GAN structure, we utilize the modified MiniMax loss from \cite{goodfellow}.The loss function is as follows.
\begin{equation}
   \mathcal{L}_{adv}=E_{(y_{gen})}[log(D(y_{gen}))]
    \label{eq:adv}
\end{equation}
where $y_{gen}$ are corresponding images taken from the original and high-resolution generated images.
\subsubsection{Final Generative loss}
Additionally to the previous losses, we add a Perceptual Loss, as they have shown incredible benefits within the training and evaluation of image generation networks \cite{johnson2016},\cite{SHI20191},\cite{liu2021},\cite{zhang2018}. Combining this with the above loss functions, we end up with the final generative loss function to be minimized by gradient descent as:
\tiny
\begin{equation}
   \mathcal{L}_{final}=\lambda_1*\mathcal{L}_{adv}+\lambda_2*\mathcal{L}_{PL}+\lambda_3*\mathcal{L}_{MGE}+\lambda_4*\mathcal{L}_{SSIM}+\\
   {\lambda_5*\mathcal{L}_{CHARB}}
    \label{eq:GENLOSS}
\end{equation}
\normalsize
with $\lambda_1$,$\lambda_2$,$\lambda_3$,$\lambda_4$, $\lambda_5$ as scaling parameters.

\subsection{Discriminative Loss Functions}
\subsubsection{Modified Minimax Loss}
In the case of discriminative losses, as the model implementation follows a GAN structure, we train the model to recognize real images as greater than 0.5, and fake images as less than 0.5. Thus, our loss is given by:
\begin{equation}
   \mathcal{L}_{disc}=E_{(x,y)}[log(D(y)]+E_{(y)}[log(1-D(G(x))]
    \label{eq:DISCLOSS}
\end{equation}

where x,y are corresponding images taken from the original and high-resolution images respectively.

\section{Experiments}
FREDSR was developed  on Tensorflow \cite{tensorflow} and Keras \cite{keras}, then trained on UHDSR4K, a SIRS benchmark dataset from \cite{Zhang_2021_ICCV}. The images were downscaled using bicubic algorithms, consistent with the original dataset.\newline\newline
Along with the techniques outlined in \textbf{Sections 1} and \textbf{2}, we utilize various traditional and novel GAN training methods, such as discriminator restart learning from \cite{Li2022} and variable Gaussian noise, an improved version of adaptive blur and control from \cite{Susmelj2017ABCGANA}. It is important to note that FREDSR utilized minimal to no hyperparameter tuning: all $\lambda$ values were set to equalize the importance of each loss function.

\subsection{Diffusive Discriminator}
A common problem with GAN stability is combating discriminator overfitting, and several methods have been used to address this, including the use of random data augmentations which adapt with the discriminator's performance. \cite{Karras2020} For stable training of GANs, we implement the solution proposed in \cite{wang2022}, using adaptive noise samples from the forward chain of a Gaussian Mixture distribution. This method has been shown both theoretically and experimentally to provide stable and data-efficient GAN training, and improve over baselines for high-quality image generation \cite{wang2022}. In this work, we implement the diffusive inputs along with our residual discriminator. 
\subsection{Discriminator Restart Learning}
Even with diffusion applied to the discriminator inputs, the discriminator tends to converge faster than the generator, or simply learn the wrong weights after an extended period of training. By resetting the discriminator learning rate mid training, or a harsh full weight reinitialization, we ensure that the discriminator continues learning based on a more converged generator output. In this reset step, the discriminator receives a learning rate increase. The generator's focus on the adversarial loss is lowered as well, to allow the discriminator to learn more freely. Once the discriminator is near convergence again, the changes to the coefficients are reversed. 
\subsection{Gaussian Random Noise}
Adding randomized noise to each convolution of the generator is a commonly used tactic in deep learning to increase the generalizability of a model while decreasing the chance of overfitting. \cite{gaussian} Although our datasets are near the larger scale, to improve the higher-resolution generalizability of our model as well as speed up training, we implement Gaussian random noise generations in between each of our fast fourier convolutions. The gaussian noise levels should decrease similarly to the generator loss, as in the later phases of training, randomized data can lead to less favorable outputs.
\subsection{AdamW Optimizer}
We utilize the AdamW optimizer proposed by \cite{Loshchilov2017} to combat the weight regularization introduced by Adam. Adam tends to regularize larger weights less than smaller weights, and thus does not converge as well as traditional methods such as stochastic gradient descent. AdamW has been proposed to improve this issue, and we notice a significant improvement over the Adam optimizer.
\subsection{Cosine Annealing Decay Learning Rate}
We utilize a cosine annealing decay learning rate scheduler to combine both decaying learning rate schedulers and warm restart learning rate schedulers proposed by \cite{Loshchilov2016}. Restart learning has been used commonly with stochastic gradient descent to speed up training processes, and exponential decay has been used to help models converge.
\begin{figure}[!ht]
    \centering
    \includegraphics[width=0.47\textwidth]{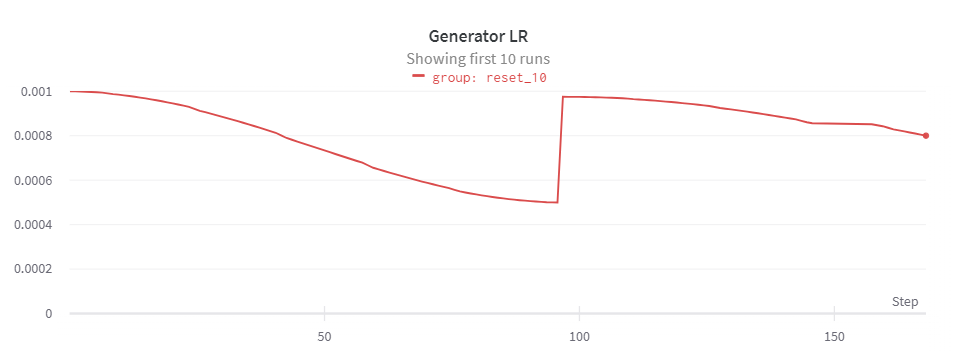}
    \caption*{Figure 4: A graph plot of the learning rate for the SISR generator: Each peak is five percent lower than the previous, and the learning rate will decay until halved prior to the next reset.}
\end{figure}
\subsection{UHDSR4K Image Super Resolution}
For our main analysis, we focus on the super-resolution objective of the UHDSR4K dataset introduced in \cite{Zhang_2021_ICCV}. This dataset is the  largest-scale UHD dataset in the field of 4K image super-resolution. For training, we downsample the original 4K images to 1920x1080. From here, we construct our 3x upsampling dataset, derived from the original UHDSR4K, by further downsampling to 640x360 resolution, and training our model to perform 3x super-resolution. This dataset is composed of diverse environments including city scenes, people, animals, buildings, cars, natural landscapes, and sculptures scraped from the internet in order to diversify the camera processing applied, along with allowing for high resolution images from many surroundings. In order to compute our downsampling, we follow \cite{Zhang_2021_ICCV} by using bicubic interpolation. For our metrics, we use standard super-resolution metrics, specifically SSIM and PSNR \cite{Wang2004},\cite{Hore2010}. Drawing comparisons using all of these methods allows for verifying that our model can both perform the super-resolution task and successfully match pixels, as well as is globally closer to the original image. It is important to note that our model's hyperparameters were not fully tuned, yielding suboptimal performance. We visualize generator loss with respect to training steps in figure 5.
\begin{figure}[!ht]
    \centering
    \includegraphics[width=0.47\textwidth]{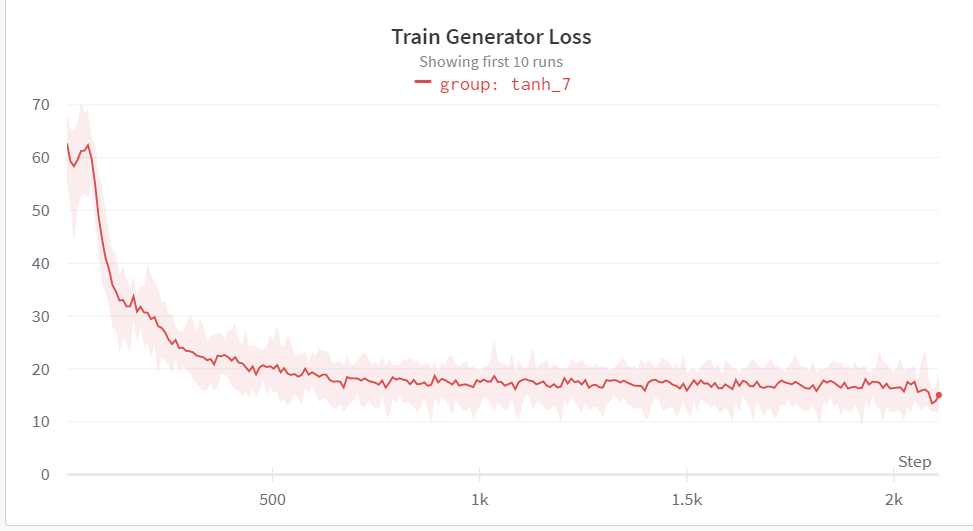}
    \caption*{Figure 5: A partial graph plot of the generator loss over train steps. Note than our model converges much slower than other models such as VDSR\cite{Kim2015}. We perform training on 12x Nvidia Tesla V100s for over 300 epochs.}
\end{figure}
\subsubsection{Comparisons to Baselines}
For our comparison, we use several baseline models, which allow for us to directly benchmark our model's performance on its original training dataset. We note that although it would be preferred to train our baseline models on the UHDSR4K dataset, as we have done with our model, due to time and resource constraints we were unable to do so. However, as this dataset has high diversity, we believe that it has coverage of the training distributions of the baseline models. We test all models at 3x super-resolution on the UHDSR4K dataset, with our training-time downscaling applied. Thus, we test 3x super resolution from 640 pixels by 360 pixels to 1920 pixels by 1080 pixels on the test dataset of UHDSR4K. For our baseline comparison models, we use only publically available pretrained models. We select 3 strong deep learning baselines with diverse techniques used. These baselines are EDSR\cite{Lim2017}, ESPCN\cite{Shi2016}, and FSRCNN \cite{dong2016}. For completeness, we also report bilinear and bicubic upscaling as a control. The results, calculated using the SSIM, and PSNR metrics as described previously are tabulated in Table 1.
\begin{table}[!ht]
\centering
\begin{tabular}{ p{2cm}|p{1cm}|p{1.1cm}|p{1cm}}

\multicolumn{4}{c}{UHDSR4K 640x320 3x Upscaling} \\
Method&Params& SSIM $\uparrow$ &PSNR$\uparrow$\\
\hline
Bicubic&0&0.856&26.75 \\
Bilinear&0&0.848&26.49 \\
EDSR&43000k&0.879&27.61 \\
ESPCN&20k&0.858&27.02 \\
FSRCNN&12k&0.857&27.03 \\
\hline
FREDSR&37k&0.883&27.776\\

\end{tabular}
\caption{Quantitative evaluation of 3x SISR performance on downscaled UHDSR4K dataset. We report Structural Similarity Index Measure(SSIM) and Peak Signal to Noise Ratio(PSNR) metrics. FREDSR compares favorably against these baselines across all metrics.}
\end{table}
As shown by this, our method can outperforms the baselines in the task of SISR on its original training resolution and dataset, strictly outperforming significantly larger models, indicating high overall performance and parameter efficiency. Further ablation analysis is necessary in order to determine the exact root of this high performance, but we believe it can be attributed to the poor generalization of the other models, high receptive field of FREDSR, and ability to take advantage of the information present in repetitive structures in order to perform more accurate prediction.
\subsubsection{High resolution Performance Across Resolutions}
\begin{figure*}[!ht]
    \centering
    \includegraphics[width=0.95\textwidth]{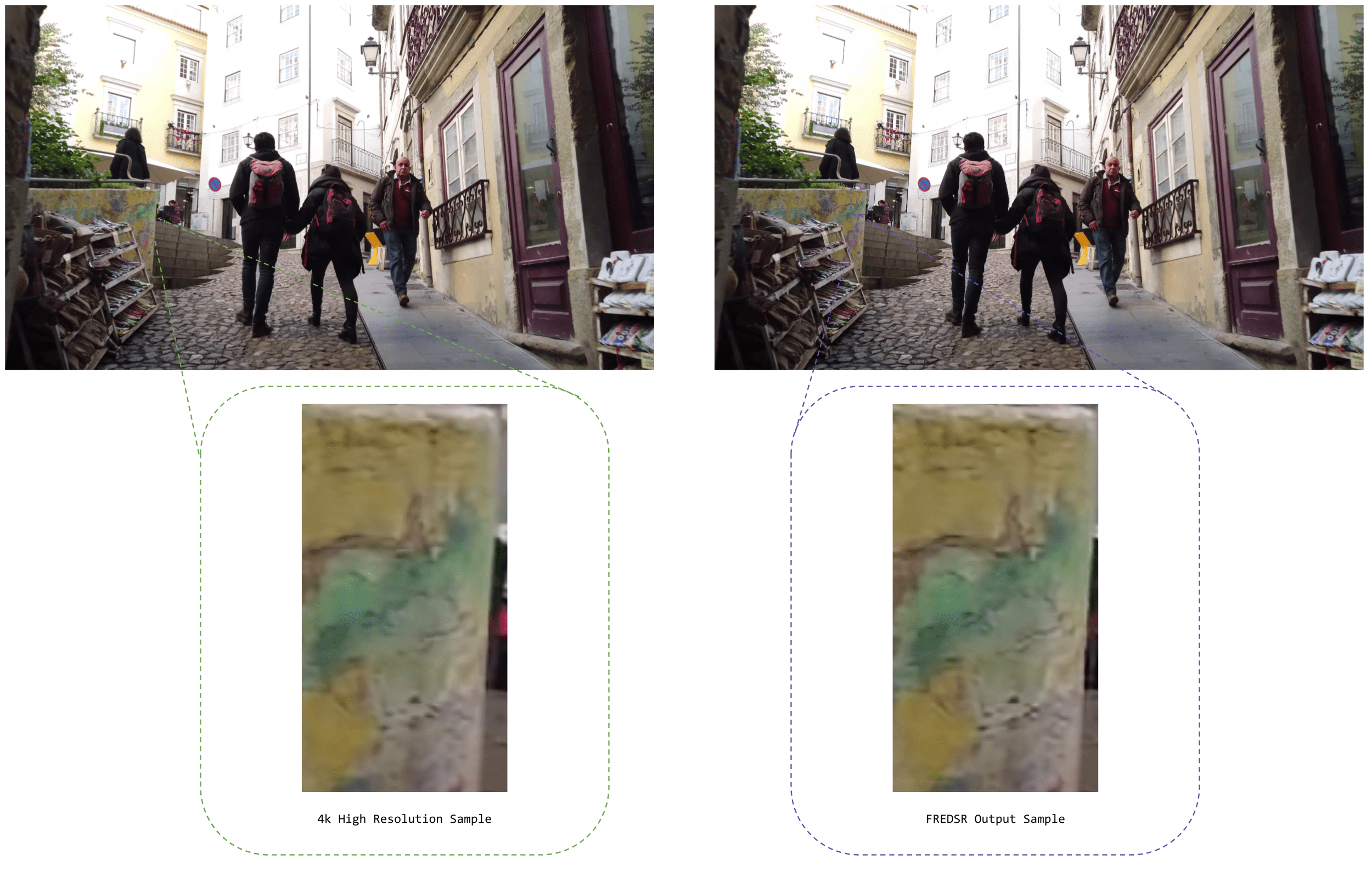}
    \caption*{Figure 6: Positive samples for high resolution performance. These are examples of our model's performance on the UHDSR4K dataset, in order to display our model's high-resolution super-resolution capabilities. The top image is our model's output, while the bottom is the true high resolution image. As shown here, along with our results from table 2, show an interesting improvement in our models as the resolution improves. This can be attributed to the task of super-resolution becoming easier at higher resolutions.}
\end{figure*}
Along with our model's performance on low resolutions, we seek to find out how it performs on high resolution upscaling. To this end, we study performance on high resolution upscaling. We point out the high amount of repetitive patches present in urban scenes, and study the ability of our network to handle repeated signals deformed by changes in perspective. This signal deformation has been shown to cause challenges to past networks implementing FFCs \cite{DBLP:journals/corr/abs-2109-07161}, thus we study our model's performance in this domain. We compare our model against a variety of strong baseline solutions for upscaling, and train a model, using the pretrained 360p model as a base, to do 3x super resolution from 720p to 4k.
\subsubsection{UHDSR4K}
For our full resolution UHDSR4K analysis, we perform 3x upscaling from the standard resolutions 720p to 4k resolutions. For our comparison, we use the PNSR and SSIM results from \cite{Zhang_2021_ICCV}, testing our models against SRCNN\cite{Dong2015}, FSRCNN \cite{dong2016},VSDR \cite{Kim2015}, EDSR\cite{Lim2017}, RCAN\cite{Zhang2018b}, RDN\cite{Zhang2018c}, HAN\cite{Niu2020}, DRLN\cite{Anwar2019}, and MANet \cite{Zhang_2021_ICCV}. We report PSNR and SSIM results from \cite{Zhang_2021_ICCV} along with our results and compare our models generalization performance at higher resolution against models that have been natively trained on upscaling the higher resolution. We also include Bilinear and Bicubic upscaling for control. 

\begin{table}[!ht]
\centering
\begin{tabular}{ p{2cm}|p{1cm}|p{1.2cm}|p{1.2cm} }

\multicolumn{4}{c}{UHDSR4K 1280x720 3x Upscaling} \\
Method&Params& SSIM $\uparrow$ &PSNR$\uparrow$ \\
\hline
Bilinear&0&0.9240&30.781 \\
Bicubic&0&0.9375&31.984 \\
SRCNN &57l&0.9503&34.082 \\
FSRCNN&12k&0.9462&33.614 \\
VDSR &665k&0.9575&35.115 \\
EDSR &43000k&0.9608&35.674   \\
RCAN &16000k&0.9608&35.576 \\
RDN &21900k&0.9614&35.769 \\
HAN &20000k&0.9601&35.547 \\
DRLN &34000k&0.9617&35.808 \\
MANet &27000k&0.9618&35.842 \\
\hline
FREDSR&37k&0.9645&35.3344\\

\end{tabular}
\caption{Quantitative evaluation of 3x SISR generalization performance on original UHDSR4K dataset. We report Structural Similarity Index Measure(SSIM), and Peak Signal to Noise Ratio(PSNR) metrics. We report metrics for other deep learning based methods from \cite{Zhang_2021_ICCV}. FREDSR compares favorably against these baselines across all metrics, even though it is comprised of far fewer parameters. This can be attributed to the scale invariance caused by the inductive biases in the FFC modules.}
\end{table}
As demonstrated here, we outperform several strong baselines, despite having orders of magnitudes of fewer parameters. This indicates an ability to efficiently capture the requisite information, which we can attribute to the use of FFC and the process of residual learning. Additionally, we see an increase in performance as resolution increases, where a model trained on multiple resolutions outperforms a model trained on only one. This is potentially due to the nature of the task of super-resolution itself, however further analysis is required to determine the root cause of this phenomenon.
\subsubsection{Other datasets}
However, we note that as expected, FREDSR does not perform as highly across other datasets, such as Manga109 or Urban100, from \cite{manga109} and \cite{urban100} respectively. This is visualized in Figure 7, where we show our poor generalization performance. The reasoning behind this is clear: bicubic upscaling does not perform as well on such datasets at all, and the learned residual images are generally lacking of color and of a much higher resolution, compared to the low resolution. This shows that FREDSR is highly specialized, losing generalizability at the benefit of extreme parameter efficiency. However, for the main real-world application of video game upscaling, this should not be a major issue, due to the consistent resolution and environment present. 
\begin{figure*}[!ht]
    \centering
    \includegraphics[width=0.95\textwidth]{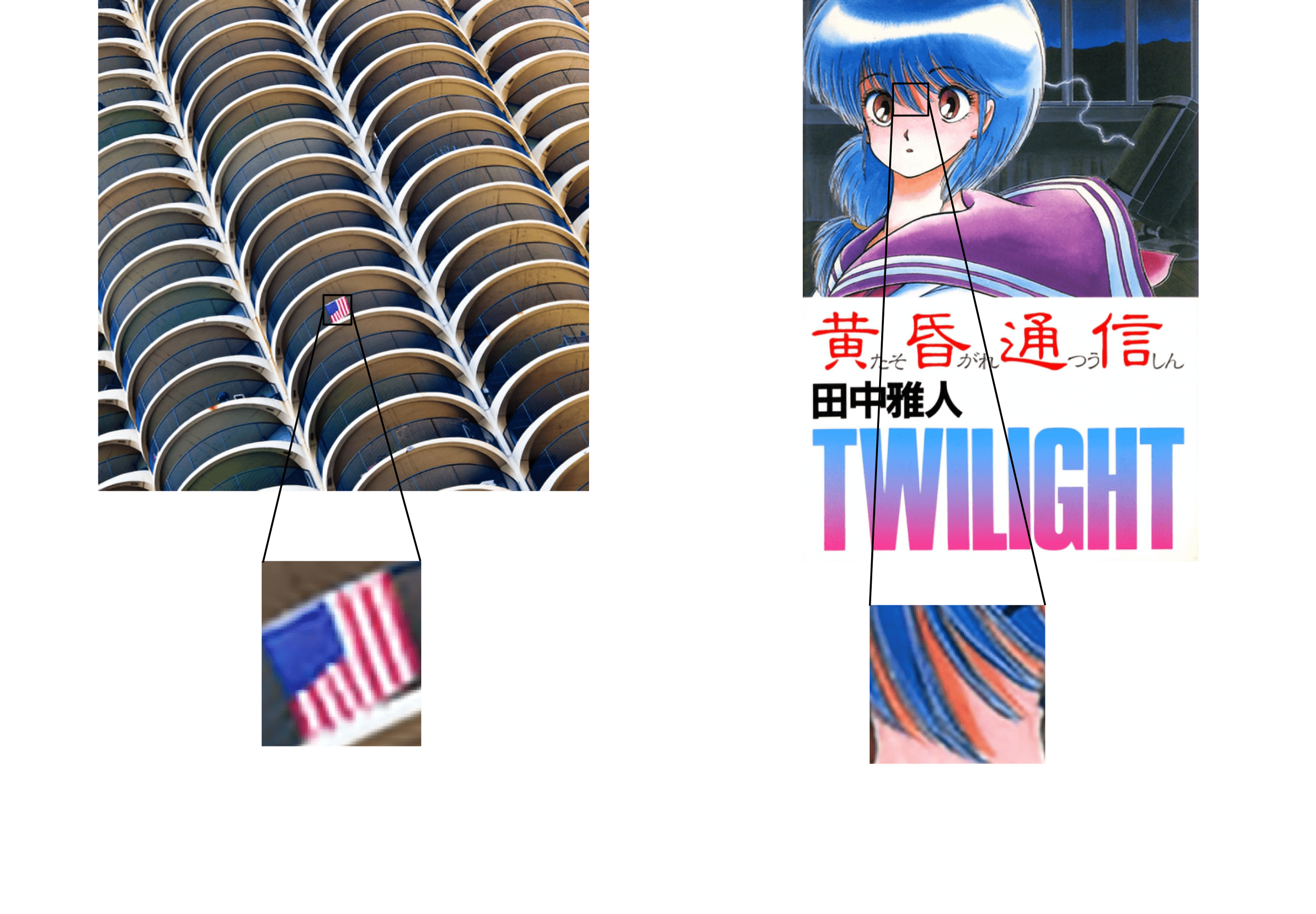}
    \caption*{Figure 7: Poor performance, with low clarity in edges, can be seen on the Manga109 or Urban100 datasets, from \cite{manga109} and \cite{urban100} respectively. This is potentially caused by the large distributional shift present, namely the change in resolution and color in residual images.}
\end{figure*}

\section{Related Work}
Within the domain of super-resolution, several traditional algorithms and data-driven models have been designed to accurately upscale images. Such classical algorithms are aptly covered in \cite{Ouwerkerk2006}, and will not be discussed further here, as most state of the art solutions to SISR are based off of learning based techniques \cite{Wang2020}. The use of convolutional networks for super resolution was first introduced in \cite{Dong2015}, which also initiated the use of deep learning for this task. Skip connections, integral to stabilizing training, improving parameter efficiency, and increasing performance \cite{Orhan2017} were first introduced with the development of DRCN \cite{Kim2015}. The use of GANs for this task, which have the major benefit of generating high-quality images when compared to more traditional methods, was introduced in \cite{Ledig2016}. The main challenge faced with GANs, their low mode coverage, or output diversity, is not applicable to the super-resolution objective, meaning that the main advantages of GANs, namely their comparatively faster sampling and high quality outputs than diffusion networks and VAEs respectively, can be utilized \cite{Xiao2022}. In a similar fashion to \cite{Ledig2016}, \cite{wang2018} combines adversarial and perceptual loss functions in order to train a GAN for super-resolution. Other works have noted the lack of high receptive field present in most super-resolution networks, pointing out that redundant patches within an image, common in both artificial and natural landscapes, allow for global context to be more informative to the computation of local upscaling. As a result, several works have taken advantage of the increased receptive field of the dilated, or atrous, convolution \cite{Seif2018}, \cite{lu2019}. This further motivates our use of Fourier convolutions, where we take advantage of their theoretically unbounded receptive fields. Other works have also used recurrent and attention mechanisms to allow for global cross-patch information usage \cite{Zhang2018b}, \cite{Niu2020}, \cite{dai2019}, \cite{lu2022}. These attention-based mechanisms are an extremely interesting novel field for further works to explore.
\section{Discussion}    
In this study, we investigate the use of the Fast Fourier convolution for solving the specialized SISR objective. Using these convolutions, we have constructed a novel GAN architecture that has shown to outperform several baselines, while being significantly smaller. Our model arguably shows good performance in urban settings due to the highly repeated sectors present, which seems challenging to other methods, as seen in figure 1 and 5. Our model seems to have extremely high parameter efficiency, outperforming models which are comprised of more than a thousand times as many parameters. In exchange for this, we lose generalizability outside of a given image distribution. We believe that this tradeoff can be harmful, but mitigated under a circumstance where we believe our method is strongest-video game upscaling. This opinion is corroborated by the fact that video games have similar highly repeating sectors to urban scenes, due to their general usage of built-in textures for three-dimensional models. This model is highly efficient, which is necessary for real-time application on consumer graphics hardware, and can be trained for a given game, as within this narrow domain significant generalization is not required. This would allow for rendering at a low resolution and upscaling. Further work would be required in this domain to allow for enforced temporal coherence.
\newline
However, convolutions are not the only approach to increasing the receptive field of image processing networks, and models like MLP Mixer models \cite{tol2020} as well as vision transformers \cite{doso2021} are exciting topics for future study in this field. Especially in the case of super resolution, higher receptive fields will unlock possibilities for the development of deep learning for high-resolution image processing, allowing for an increase in fidelity, diversity, and processing speed. Improved super resolution will allow for advancements in medical imaging \cite{shi2013} and compression of image and video formats \cite{cao2020}, among many other applications.
\section*{Acknowledgments}
The authors acknowledge the MIT SuperCloud and Lincoln Laboratory Supercomputing Center for providing (HPC, database, consultation) resources that have contributed to the research results reported within this paper. We additionally thank Professor Leslie Kaelbling and Ge Yang from the MIT Computer Science and Artificial Intelligence Laboratory for their valuable discussions and providing us with access to resources to complete this research. Lastly, we acknowledge Tensorflow and Horovod as the development language and platform of choice for this project.
\newpage
\onecolumn
\bibliography{refs.bib} % Entries are in the refs.bib file

\end{document}